\pdfoutput=1
\documentclass[11pt]{article}

\usepackage{ACL2023}

\usepackage{times}
\usepackage{latexsym}
\usepackage{graphicx}
\usepackage{caption}
\usepackage{adjustbox}
\usepackage[T1]{fontenc}
\usepackage[utf8]{inputenc}

\usepackage{microtype}

\usepackage{inconsolata}

\usepackage{multicol}
\usepackage{hyperref}

\newcommand{\tb}[1]{\textcolor{blue}{#1}}

\newcommand{\dataset}{FNXL}
\newcommand{\modelname}{AttentionXML Pipeline}
\newcommand{\axml}{AttentionXML}
\newcommand{\finer}{FiNER}

\title{Financial Numeric Extreme  Labelling: A Dataset and Benchmarking for XBRL Tagging}

\author{
\renewcommand*{\thefootnote}{\fnsymbol{footnote}}
\textbf{Soumya Sharma}$^{\spadesuit}$ *~~~~
\textbf{Subhendu Khatuya}$^{\spadesuit}$ *~~~~
\textbf{Manjunath Hegde}$^{\diamondsuit}$ ~~~~
\textbf{Afreen Shaikh}$^\diamondsuit$\\
\textbf{Koustuv Dasgupta} $^{\diamondsuit}$ ~~~~
\textbf{Pawan Goyal}$^{\spadesuit}$ ~~~~
\textbf{Niloy Ganguly}$^{\spadesuit}$\\ 
\\
$^\spadesuit$Indian Institute of Technology, Kharagpur \\
$^{\diamondsuit}$Goldman Sachs, Data Science \& Machine Learning ~
}

\begin{document}
\maketitle
\begin{abstract}
The U.S. Securities and Exchange Commission (SEC) mandates all public companies to file periodic financial statements that should contain numerals annotated with a particular label from a taxonomy. In this paper, we formulate the task of automating the assignment of a label to a particular numeral span in a sentence from an extremely large label set. Towards this task, we release a dataset, Financial Numeric Extreme Labelling (\dataset), annotated with 2,794 labels. We benchmark the performance of the \dataset\ dataset by formulating the task as (a) a sequence labelling problem and (b) a pipeline with span extraction followed by Extreme Classification. Although the two approaches perform comparably, the {\sl pipeline solution} provides a slight edge for the least frequent labels. %We obtained higher performance using the pipeline solution than the NER paradigm, however, there are some cases where NER methods perform better. %\pg{This is not so convincing, need to revisit.}
\end{abstract}

\def\thefootnote{*}\footnotetext{These authors contributed equally to this work}\def\thefootnote{\arabic{footnote}}

\section{Introduction}

%\st{The U.S. Securities and Exchange Commission (SEC) mandates public companies to file periodic financial statement and other disclosures. These documents are of importance to finance professionals and investors who rely on SEC filings to make informed decisions. There are multiple type of filings that a company is required to submit including 10-K (annual reports), 10-Q (quarterly reports) etc. }
In 2019, the U.S. Securities and Exchange Commission (SEC) mandated each company to use GAAP  metrics\footnote{GAAP: Generally Accepted Accounting Principles, the number of metrics is continuously evolving. In 2021, the total number of metrics was 20,323. About 6K in textual content.} to standardise financial reporting. These metrics are used to tag portions of SEC documents including, numerals using eXtensive Business Reporting Language (XBRL), an XML based language to facilitate the processing of financial information. 
%The taxonomy is an ever evolving list of metrics provided by SEC. In 2021, the total number of metrics stood at 20323. 
The process of annotating the documents requires enormous manual effort: expert annotators from a company have to go through the document and mark each relevant detail with a relevant GAAP %\pg{name this consistently} \soumya{Resolved}
metric label.  
This necessitates the development of an automatic annotation process that may reduce the manual effort to annotate the documents. Solving this task would also help with annotation of old as well as new reports (which may not contain XBRL tags). Towards this goal, we aim to decrease the list of possible tags for annotators and provide them with a crisp list of k tags.

%It is also found that due to the manual effort involved, most companies only tend to annotate a few sections of the entire report in detail \noteng{reference}, hence the automation may help in annotating the left-out parts. Finally there are a large number of historically un-annotated documents which can be annotated and made compatible for state-of-the-art (tool) analysis. %, (b) annotating the unannotated documents (old and new), (c) annotating the unannotated parts of the documents.

\begin{figure}[h]
    \centering
    \includegraphics[scale=0.6]{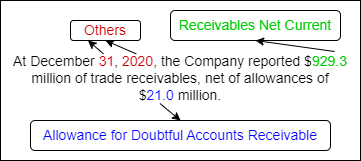}
    \caption{An annotated example from \dataset~ dataset.}
    \label{fig:example}
    %\vspace{-6mm}
\end{figure}

%In this work we tackle the automation problem, however focusing  only on numericals found in SEC filings annotated with a US-GAAP metric label. 
We define the problem as, given a sentence, identify the relevant numerals and assign them a particular GAAP label. %For example, for a sentence, ``At December 31, 2020, the Company reported \$929.3 million of trade receivables, net of allowances of \$21.0 million.'', the task is to identify the numerals and label them with their appropriate label. In this sentence, the numeral $\textbf{929.3}$ is tagged as \textit{``Receivables Net Current''} and $\textbf{21.0}$ is tagged as \textit{``Allowance For Doubtful Accounts Receivable''}.
An example of this annotation is provided in Figure ~\ref{fig:example} whereby each numeral is marked with a label or identified as `other'.

%\ which released a similar dataset with 139 labels (chosen according to most frequently occurring). \finer\ solved 

% \finer\ paper 
This problem has previously been tackled by \finer ~\cite{loukas2022finer}, 
as a sequence labelling approach using BERT ~\cite{devlin2019bert}, where their set of labels is the 139 labels that occur the most frequently. 
However, we find that in the real-world setting, a much larger number of labels are used to annotate the text found in these documents. 
Thus, to evaluate the real world context, we painstakingly prepare a dataset, \dataset\ that contains a total of 2,794 tags. 
%enabling us to evaluate the real-world setting. 
We realize that given the sheer number of labels, extreme classification is an ideal methodology that can be followed. 
Subsequently, we find that the \axml\ method ~\cite{you2019attentionxml} from the Extreme Classification (XC) repository ~\cite{Bhatia16} can be modified to suit our problem setting. 
Consequently,  we use the dataset, \dataset\, to benchmark both the \finer\ based sequence labelling approach and the \axml\ method. 
We find that XC methods give a comparable performance (better in certain situations) as \finer. Specifically, XC performs better in tagging infrequent labels. 
%in Macro values, \finer\ performs better in Micro values. We also find that \axml\ pipeline performs better at assigning tail labels (less training data) when compared to \finer.}

%We expand on this in the upcoming sections and make the following contributions:
%\begin{enumerate}
%    \item We release a corpus of SEC filings spanning from 2019, 2020 and 2021.
%    \item We use a subset of the corpus to create a dataset containing  sentences and  annotated numericals.
%    \item We benchmark the dataset by adapting different methods from the Extreme Classification Repository for our purpose and through the \finer method.
%\end{enumerate}

% \pg{See the flow from Abstract. Need to talk about FiNER's small label set, which is not the realistic picture.}
% Keeping the problem and the large number of metrics in mind, we take inspiration from Extreme Classification Repository and formulate the problem. However, unlike the XMC task which attempts to classify an entire document with a set of labels, our task is to assign a label (GAAP metric) to a particular numerical present in a sentence. To this purpose, we release the \dataset dataset containing 79088 sentences and a total of 2930 labels. From these sentences, a total of 147281 numericals contain a label. This dataset has been created from filings of 2339 companies spanning the years 2019-2021. 

%\input{Sections/ProblemFormulation}
\section{Dataset Description}

\subsection{Data Sources}
To promote transparency, the government body, SEC (U.S. Securities And
Exchange Commission) mandates publicly reporting companies to publish reports in order to disclose information at various intervals. For example, domestic companies must submit annual reports on Form 10-K, quarterly reports on Form 10-Q, and current reports on Form 8-K for a number of specified events, and must comply with a variety of other disclosure requirements. We use the publicly available\footnote{\href{https://www.sec.gov/}{https://www.sec.gov/}} annual 10-K reports from 2019-2021 for 2,339 companies as the source of our \dataset\space dataset. Annual reports are mandated by SEC to be annotated using XBRL (eXtensible Business Reporting Language) which is a freely available and global framework for exchanging business information. XBRL contains an ontology of metrics that include the GAAP: Generally Accepted Accounting Principles metrics. The number of metrics are continuously evolving and in 2021, the total number of metrics stood at 20,323. Out of these, about 6K were found in textual content.

Every annual report\footnote{\href{https://bit.ly/example-10k}{https://bit.ly/example-10k}} contains 4 parts and 15 schedules out of which typically only 3 schedules contain XBRL metric annotated data: (a) Item 7. {\small \textsc{Management's Discussion and Analysis of Financial Condition and Results of Operations.}} (b) Item 8. {\small \textsc{Financial Statements and Supplementary Data}}, and (c) Item 9. {\small \textsc{Changes in and Disagreements with Accountants on Accounting and Financial Disclosure.}}

\subsection{FNXL Dataset}

In this work, we focus only on annotated textual data and discard annotated tables. We analyse the fillings for the 2,339 companies and find that 160K sentences are annotated across these fillings, we %and preprocess the data obtained including filtering 
filter out sentences with less than 50 characters and  annotated data that is alphabetic; we only retain \underline{numeric annotated data}. We also perform some manual cleaning on the dataset to remove some noisy datapoints. However, %. SEC documents contain labels which are obtained from particular taxonomies. eg: SEC defines the US-GAAP taxonomy while 
companies might annotate some numerals in the document with a self-defined taxonomy; consequently, we filter out all labels that are not US-GAAP labels. 
Finally, our Financial Numerical Extreme Labelling (\dataset) dataset\footnote{Code and Dataset available at: \href{https://github.com/soummyaah/FNXL}{Github Link}},  contains a total of 79,088 sentences containing 142,922 annotated numerals with a label set of size 2,794. 

\begin{table}[t]
\begin{center}
\resizebox{\columnwidth}{!}{
\begin{tabular}{||c | c c c c||} 
 \hline
 - & \# sentences & \# companies & \# data points & \# labels \\
 \hline\hline
 Train & 62,782 & 798 & 111,493 & 2,692 \\ 
 \hline
 Dev & 6,823 & 756 & 13,191 & 1,273 \\
 \hline
 Test & 9,483 & 794 & 18,238 & 1,374 \\
 \hline
 Total & 79,088 & 840 & 142,922 & 2,794  \\
 \hline
\end{tabular}}
\end{center}
\caption{Train-test-dev division of datapoints. A datapoint is a numeral tagged with US-GAAP metrics, multiple datapoints may occur in a single sentence.}
\label{table:data_stats}
\end{table}

To avoid data leakage, we divide the sentences according to the companies they belong to and create the train, validation and test set. We ensure that the companies in train set do not contribute to the validation or test set and vice-versa. This results in a approximately 78:9:13 percentage division between the train, validation and test set. We present the exact numbers in Table \ref{table:data_stats}. The sentences have an average length of 37.83 tokens, stdev of 20.37 tokens and a maximum length of 590 tokens.

The validation and the test set also contain 40 and 69 labels not seen in the training data corresponding to 76 and 119 numerals, respectively. Similar to the entire dataset, these zero-shot data points also come from unseen documents and unseen companies in train set. % We add sentences corresponding to all new labels from 9 and 13 companies in the validation and test set respectively. To ensure that the model can predict the label on unseen documents and for unseen companies, we add sentences obtained from 9 and 13 unseen companies in the validation and test set respectively. These sentences contain a total of 1,037 and 1,430 annotated numerals in the validation and test set respectively. %\pg{At one place, you mention `data points', another place, `numerals', please be consistent.} \soumya{Modified.} 

\subsection{Label Set Details}

\begin{figure}[h]
    \centering
    \includegraphics[width=\columnwidth]{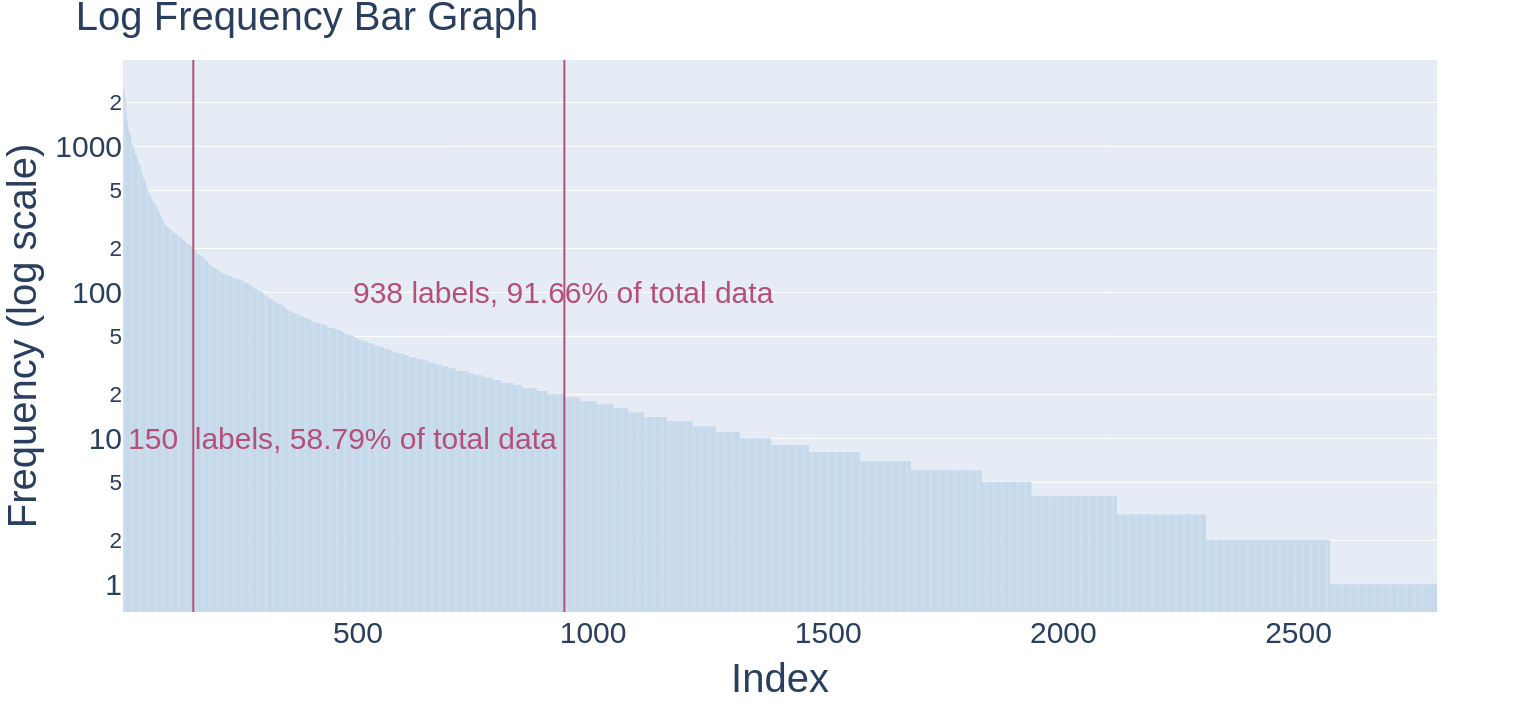}
    \caption{Scatter-plot showing the log frequency distribution of the 2,794 labels in train set of \dataset. The ordering is from highest to lowest frequency.}
    \label{fig:frequency}
    %\vspace{-6mm}
\end{figure}

While in the FiNER-139 dataset ~\cite{loukas2022finer}, only the 139 most frequent XBRL tags with at least 1,000 appearances in the dataset are selected, we keep our data label set unfiltered and obtain a set of 2,794 labels. We find that 100 labels from the FiNER-139 dataset are part of our label set. We showcase the frequency distribution of our dataset in Figure \ref{fig:frequency}.

%\textcolor{red}{TODO: Add plot here.} We find that the frequency distribution is a long-tailed distribution. \textcolor{red}{TODO: Add which plot.}

\begin{table}[h]
\begin{center}
\resizebox{\columnwidth}{!}{
\begin{tabular}{|c|ccc|}
\hline
                                  & Max  & Min & Avg ($\pm$ Std dev) \\ \hline
Data points per label             & 2,529 & 1   & 51.15 ($\pm$ 168.07) \\
Label density per sentence        & 17   & 0   & 1.81 ($\pm$ 1.04) \\
Unique label density per sentence & 8 & 0 & 1.18 ($\pm$ 0.51) \\
Number of tokens in label name         & 23   & 1   & 7.67 ($\pm$ 3.79) \\ \hline
\end{tabular}}
\end{center}
\caption{Some statistics around the labels in the \dataset\space dataset.}
\label{table:label_stats}
\end{table}

In our \dataset\ dataset, we see that the top 150 frequently occurring labels (each containing more than 200 data points) out of 2,794 correspond to 58.79\% of our total data points and the least 1,856 frequently occurring labels (each containing less than 20 data points) constitute 8.34\% of our total data points. Some label specific statistics are given in Table \ref{table:label_stats}.
% The maximum number of data points per label is at 2529, the minimum is 1, and the average is 51.15 with a standard deviation of 168.07.} %, thereby, showcasing the need for such a real-world dataset. % \soumya{modified}.
%how many labels are most frequently occuring and least frequently occurring. 

%\subsection{Label Set statistics}
%We obtain 2930 labels in total. Each of these labels are a part of the US GAAP Taxonomy. In table ~\ref{table:label_stats} we showcase how many labels %are most frequently occuring and least frequently occurring. 

We also analyse the cosine similarity of BERT representations of the names of the labels. We find that the average cosine similarity for the 5th most similar tag is 71.73\% and maximum is 99.02\%.
% We find that top@1 cosine similarity is an average of  86.62\% and maximum of 99.76\%. Top@3 cosine similarity is an average of  76.47\% and maximum of 99.32\%. 
% We find that Top@5 cosine similarity is an average of  71.73\% and maximum of 99.02\%. This showcases that the similarity between the names of the labels is very close till atleast Top@5. 
One example of this is where the label "Other Comprehensive Income Loss Derivative Excluded Component Increase Decrease Before Adjustments After Tax" is very close to "Other ... Decrease After Adjustments Before Tax
", "Other ... Decrease Adjustments After Tax", "Other ... Decrease Before Adjustments Tax".
%We report that the per sentence label density values are on average 1.77, maximum is 17 and minimum is 1. The maximum number of tokens in the labels are 23, minimum is 1 and average is 7.67 tokens on average. We also look at how close the labels are. We analyse the cosine similarity of the names of the labels and we find that \textcolor{red}{TODO: Finish this part.}}

% In Figure ~\ref{fig:frequency}, we showcase the frequency of the labels vs the label index. As can be seen from the long tail, our dataset contains many labels where data points < 10. 

%\subsection{Data Statistics}

\section{Benchmarking}

We extensively benchmark the dataset using two broad class of algorithms, the named-entity based \finer\ model and Extreme Classification (XC) based \axml\ model. We briefly describe these models and the different variations tried.
%\pg{Motivate why these two.}

\subsection{FiNER}
The Fine-Grained Named Entity Recognition (FiNER) approaches the task %\pg{Is this a classification problem? You are creating a pipeline to treat it as classification.} \soumya{Modified.} 
as a named entity recognition task, aiming to assign a label to each subword within a sentence. %This is achieved by fine-tuning BERT-BASE to generate contextualized embeddings of the subwords, which are then fed into a multinomial logistic regression layer for prediction of the corresponding subword's label. Overall, FiNER 
\finer\ utilizes contextualized subword embeddings from fine-tuned BERT and a logistic regression layer to accurately classify named entities in text. %\pg{No mention of BERT?}\soumya{Modified.}
\finer\ identified an issue with over-fragmentation of numerals by the BERT tokenizer, which negatively impacts the performance of subword-based models. To overcome this problem, \finer\ introduced two pseudo-tokens: [NUM], representing an entire numeral token, and [SHAPE], representing the shape of a numeral (e.g., 54.3 would be replaced by [XX.X]). % and 78,231.9 would be replaced by [XX,XXX.X]). 
They also additionally release SEC-BERT based models %, SEC-BERT-NUM, SEC-BERT-SHAPE models, 
which are BERT-BASE models pre-trained on the EDGAR-CORPUS ~\cite{loukas2021edgar}. 
%and incorporate [NUM] and [SHAPE] psuedo-tokens, respectively. Overall, the use of these pseudo-tokens and pre-trained models allows \finer\ to effectively handle numerals in the named entity recognition process.

%We experiment with the masking technique introduced by \finer\ and 
%use it with BERT-BASE and SEC models. In total, we 
We showcase the results for six \finer\ based models: three each of BERT-BASE and  SEC-BASE, respectively. For each, the three models are {no-masking, [NUM] and [SHAPE]}. %  {no-masking, [NUM], [SHAPE]}. 
%, SEC-NUM, SEC-SHAPE where SEC-NUM and SEC-SHAPE use the [NUM] and [SHAPE] masking technique respectively.
%\pg{You can end this section describing in brief as to which of these models you have selected.} \soumya{Modified.}

\subsection{\axml\\}
Extreme Classification (XC) methods have shown to be effective on real-world datasets where the distribution of data points is extremely skewed and many tail labels often have very few data points to be trained on. Due to the similarity of our dataset with the XC datasets, we adapt an XC method to our use-case. In particular, we benchmark using \axml\ which focuses on the entire input and not a particular span; we use a two-step approach to identify and label numerals in a sentence. %\pg{Not a great opening sentence -- talk about it being used for extreme classification, and not being directly applicable to our problem definition. Motivate the need for the pipeline and then discuss this} \soumya{Modified.} 
The first step is to identify the relevant numerals in a sentence and the second step is to label the numerals with their corresponding label. We describe the two steps in detail below.

\noindent{\bf Binary Classifier:} 
We use a BERT-based sequence tagger to identify the relevant numerals in a sentence. This tagger marks each numeral of a sentence with a label indicating whether it is a relevant numeral or not. %The bert-base-uncased transformer model is fine-tuned to extract the contextualized embeddings of subwords, and a multinomial logistic regression layer is applied on top of these embeddings to predict the relevancy of a numeral.

%\subsection{Numeral tagger} To tag a particular numeral with a label, we adapt the AttentionXML model. AttentionXML attempts to label an entire input instead of labelling a particular span.
% We adapt the AttentionXML model and through variations in input focus on a particular span. The module takes as input a particular sentence with the target numeral marked in a special way. We make the following changes to AttentionXML:
% 1) We replace the Glove + BiLSTM layer of the Attention-Aware Deep Model with BERT representations of the input. 
% 2) We apply a local attention layer after obtaining the BERT representations. The local attention layer attends to the span of the target numeral.
% 3) We also experiment with the concept introduced by \finer to avoiding defragmentation and replace numerals with the [NUM], [SHAPE] token. Here, the target numeral is left untouched and the others are replaced by the token. We showcase our masking strategies in Figure ~\ref{fig:token_replacement_eg}.

\begin{figure}[h]
    \centering
    \includegraphics[scale=0.50]{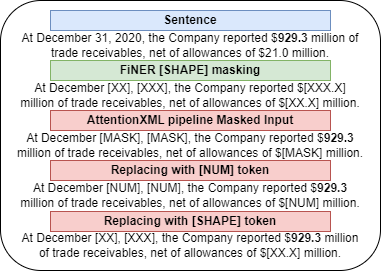}
    \caption{The target numeral is 929.3, and the masking strategies for \finer\ and \axml\ are shown.}
    \label{fig:token_replacement_eg}
    %\vspace{-6mm}
\end{figure}

%Using the above mentioned variations, we experiment with AttentionXML by using different masking strategies including unmasked, masking non-relevant tokens with [MASK], masking non-relevant tokens with [NUM], masking non-relevant tokens with their shape token, example: [XX.XXX] for 33.541.

\noindent{\bf AttentionXML Model:} It is a deep learning approach for multi-label text classification, consisting of five layers: a word representation layer, a bidirectional Long Short-Term Memory (BiLSTM) layer, a multi-label attention layer, a fully connected layer, and an output layer. GloVe word embeddings of tokenized text is fed as input to the the BiLSTM layer. The output is then passed through a multi-label attention mechanism introduced in ~\citet{you2019attentionxml}, which allows to capture the important parts of texts most relevant to each label. The model also includes one or two fully connected layers and an output layer, with shared parameters across all labels to reduce the risk of overfitting and keep the model scale small. The binary cross-entropy loss function is used to train the model. %\pg{In this whole description, there is nothing specific about attention XML -- hierarchy is not even talked about, which is the central point. Others are anyway used in basic architectures.}\soumya{Modified.}

While \finer\ uses masking strategies to mask all numerals in a sentence, here we only mask the irrelevant numerals to psuedo-focus on the relevant numeral span. A relevant numeral here means the numeral for which we want the model to assign a label. We experiment with different masking strategies such as [MASK], [NUM] and [SHAPE] and showcase an example in Figure ~\ref{fig:token_replacement_eg}.

\subsection{Evaluation Metrics}
To evaluate the setting fairly we use the following metrics: 1) Macro-Precision 2) Macro-Recall 3) Macro-F1 4) Micro-Precision 5) Micro-Recall 6) Micro-F1. 
The macro-averaged F1 score is computed using the arithmetic mean of all the per-class F1 scores. For financial numeral labelling, all the tags are equally important. So using the macro average is a good choice as it treats all classes equally regardless of their frequency. 
\section{Results}

\begin{table*}[ht]
\begin{center}
% \resizebox{\columnwidth}{!}{
{\small{
\resizebox{\linewidth}{!}{
\begin{tabular}{| c | c | c | c | c | c | c | c |}
\hline Model & Masking Token & Macro-Precision & Macro-Recall & Macro-F1 & Micro-Precision & Micro-Recall & Micro-F1\\
\hline
\finer\ (BERT-base) & no-masking & \tb{\textbf{49.17}}	& \textbf{49.71}	& \tb{\textbf{47.13}} & 76.493 & \tb{\textbf{75.21}} & \textbf{75.84}   \\ \hline
\finer\ (BERT-base) & [NUM] &  48.86         &    48.01    & 46.16 & \textbf{76.51} & 74.68 & 75.58\\ \hline
\finer\ (BERT-base) & [SHAPE] &    42.74       &   43.93     &  40.62 & 72.13 & 72.56 & 72.35\\ \hline\hline
\finer\ (SEC-base) & no-masking & \tb{\textbf{47.76}}	& \tb{\textbf{48.87}} &	\tb{\textbf{46.20}} & \tb{\textbf{75.84}} & \textbf{75.84} & \tb{\textbf{75.84}}\\ \hline
\finer\ (SEC-num) & [NUM] &      44.62    &    45.80    & 42.74 & 74.32 & 74.59 & 74.45\\ \hline
\finer\ (SEC-shape) & [SHAPE] &     45.53      &     45.34   &  42.93 & 75.15 & 73.33 & 74.23 \\ \hline\hline
\modelname & [MASK] & 49.83           & 47.99        & 46.58  & 73.91 & 74.37 & 74.14  \\ \hline
\modelname & [NUM] & 49.01           & 48.25        & 46.49 & 73.57 & 74.03 & 73.8\\ \hline
\modelname & [SHAPE] & \textbf{50.69}           & \tb{\textbf{48.51}}        & \textbf{47.54} & \tb{\textbf{74.5}} & \tb{\textbf{74.96}} & \tb{\textbf{74.74}}\\ \hline

\end{tabular}
}
\caption{\label{table:result_table}Performance evaluation based on Macro and Micro metrics by \finer\ and \modelname\ }
}}
\end{center}
\end{table*}

We report the results for 9 experiments in total, 6 \finer\ based models and 3 \axml\ based models as showcased in Table~\ref{table:result_table}. 

We observe that the \axml\ pipeline performs  better in Macro scores than the \finer\ model. The best performing \axml\ pipeline uses the [SHAPE] masking token and achieves a 47.54\% Macro-F1 vis-a-vis 47.13\% Macro-F1 for best the \finer\ model with no-masking. However, \finer\ performs better in the Micro values achieving a 75.84\%  Micro-F1 vis-a-vis 74.74\% Micro-F1 for \axml\ pipeline.
The superior performance of \finer\ in Micro metrics can be attributed  to its better performance with popular labels as the frequency distribution of labels follow a long-tail distribution. This is elaborated in  Section ~\ref{sec:bucket_analysis}.

We also observe that the masking technique does not help the \finer\ model as can be seen in the case of BERT-base \finer\ models. 
%where the no-masking token model performs best with a Macro-F1 of 47.13\% and the case of SEC models where SEC-BASE performs best with a Macro-F1 of 46.20\%. 
We note that the BERT-base models perform better than the SEC based models in terms of Macro-F1 and comparable in terms of Micro-F1. %While the masking token trend is observed in the Micro scores as well, the SEC-BASE Micro-F1 is comparable to BERT-BASE Micro-F1. 
The adapted masking technique is beneficial in case of \axml\ pipeline. We theorize that since \axml\ uses GloVe word embeddings, it does not face the problem of defragmentation.  Also, unlike \finer, in \axml\ masking is done on irrelevant numerals which helps the model focus on the context of the relevant numeral. 
From the results we can establish that the [NUM] and [SHAPE] psuedo-tokens help \axml\ model successfully generalize over numeric expressions. 
%With [SHAPE], we get best results in almost all scenarios. Since the shape pseudo-tokens capture the magnitude of the numeric, we theorize that tokens of similar magnitudes may require similar xbrl tags, thereby, giving the performance of [SHAPE] based model a boost.

%Our pipeline achieves reasonable performance across all of the performance metric. Our pipieline with shape masking strategy achieves best result in terms of macro-precision and macro-f1.

 %\pg{No detail of results? Or are you waiting for the final numbers before you add this?}

 %\pg{Naming of models across tables is not consistent. See Table 2 vs Table 5}
%\soumya{Modified.}
 
 %We also experimented with BERT, FinBERT and SEC-BERT instead of Glove in the AttentionXML framework, but that performed poorly. SEC-BERT with [NUM] and [SHAPE] performed better than BERT and FinBERT.  FinBERT performs slightly better than BERT as its pre-trained on financial data.

\subsection{Bucket Analysis}\label{sec:bucket_analysis}
%To understand the nature of difference in performance between \finer\ and \axml\, 
%We  compare the performance of \finer\ and \axml\ pipeline with respect to the top and least frequently occurring labels.
%and least frequently occurring labels. Due to space shortage and low over-all performance, we do not analyse SEC \finer\ models.
Table ~\ref{table:bucket analysis} shows   the average performance across \finer\ and \axml\ pipeline models for top-100 and bottom-1000 frequent occurring classes (for both we consider the best performing model.  
%which constitute 50.65\% of the data points across the dataset. Here, we report the average across \finer\ and \axml\ pipeline models. 
Exhaustive model wise results are shown in Appendix ~\ref{sec:appendix}). 
The performance of \finer\ is superior for frequently occurring labels while it is vice-versa  for infrequent tokens. This confirms the reason behind \axml\ pipelines' better performance in the Macro related metric where each class (label) is giving the same importance. Due to space constraints, we give the top frequently occurring label analysis in Appendix ~\ref{sec:appendix}.

\begin{table}[ht]
\begin{center}
\resizebox{\columnwidth}{!}{
\begin{tabular}{|c|c|c|c|c|}
\hline
                                Model & Masking Token & Macro-Precision & Macro-Recall & Macro-F1       \\ \hline
FiNER (BERT-base) & no-masking              &  43.28               &       38.88       &      40.03          \\ \hline
FiNER (BERT-base) & [NUM]           &    \textbf{45.54}             &     \textbf{40.24}         &    \textbf{41.76}            \\ \hline
FiNER (BERT-base) & [SHAPE]         &   38.97              &       34.22       &   35.70             \\ \hline
FiNER avg. &  & 42.601	& 37.783 &	39.166 \\ \hline\hline
% FiNER (sec-base)                &       48.67          &       43.23       &         44.77       \\ \hline
% FiNER (sec-num)                 &        40.42         &       37.05       &      37.60          \\ \hline
% FiNER (sec-shape)               &     38.56            &       35.23       &       36.24         \\ \hline
% \modelname + unmasked & 32.55           & 29.07        & 29.81          \\ \hline
\modelname\ & [MASK]   & 45.33           & 40.79        & 42.12          \\ \hline
\modelname\ & [NUM]      & \textbf{45.87}           & \textbf{41.48}       & \textbf{42.77}          \\ \hline
\modelname\ & [SHAPE]    & 45.33  &    40.44        & 41.83 \\ \hline
\modelname\ avg. &  & \tb{\textbf{45.513}} &	\tb{\textbf{40.91}} &	\tb{\textbf{42.245}} \\ \hline
\end{tabular}}
\caption{\label{table:table_least_1000}Least 1000 frequent occurring class}
\end{center}
\end{table}

\textbf{Least frequently occurring labels:} We observe that that masking with [NUM] token provides the best performance for both the models with \finer\ with [NUM] token masking showcasing a 41.76\% Macro-F1 and \axml\ pipeline with [NUM] token masking showcasing a 42.77\% Macro-F1. On average, \axml\ pipeline performs better than \finer.

%We observe that \finer\ models perform better with a Macro-F1 score of 82.52\% whereas \axml pipeline achieves a Macro-F1 score of 81.97\%. Next, we analyzed the result considering least-1000 frequent occurring classes (Table \ref{table:bucket analysis}). \soumya{TODO: List the number and \% of data} In this case, due to very limited examples for each class the performance drops for \finer\ and \axml\ shows a relative improvement of 7.86\% on comparing average values. We analyse that due to this jump in performance for least frequently occurring classes, there is a difference in overall model performance and \axml\ performs better in macro results whereas \finer\ performs better when comparing micro results. 

% In this case, FiNER  with sec-base variation achieves best macro recall and macro f1, whereas sec-num version provides best precision. Except unmasking one, all other variations of the models give more than 80\% macro-f1 score in this bucket. For FinTag, We get best performance with shape strategy for this bucket.

% begin{center}
% \resizebox{\columnwidth}{!}{
% \begin{tabular}{||c | c c c c||} 
%  \hline
%  - & \# sentences & \# companies & \# data points & \# labels \\
%  \hline\hline
%  Train & 62782 & 789 & 114704 & 2695 \\ 
%  \hline
%  Dev & 6823 & 756 & 13692 & 1285 \\
%  \hline
%  Test & 9483 & 794 & 18882 & 1389 \\
%  \hline
%  Total & 79088 & 2339 & 147281 & 2930  \\
%  \hline
% \end{tabular}}
% \label{table:data_stats}
% \end{center}

\begin{table}[ht]
\begin{center}
\resizebox{\columnwidth}{!}{
\begin{tabular}{|c|c|c|c|}
\hline
\multicolumn{1}{|l|}{Model}                      & \multicolumn{1}{l|}{Macro-Precision} & \multicolumn{1}{l|}{Macro-Recall}   & Macro-F1       \\ \hline
\multicolumn{4}{|l|}{\textit{Top 100 frequently occurring labels}} \\ \hline
\multicolumn{1}{|l|}{FiNER avg.}                 & \multicolumn{1}{l|}{\tb{\textbf{90.28}}}  & \multicolumn{1}{l|}{\tb{\textbf{77.94}}} & \tb{\textbf{82.52}} \\ \hline
\multicolumn{1}{|l|}{AttentionXML Pipeline avg.} & \multicolumn{1}{l|}{88.81}           & \multicolumn{1}{l|}{77.87}          & 81.97          \\ \hline
\multicolumn{4}{|l|}{\textit{Least 1000 frequent labels}} \\ \hline
\multicolumn{1}{|l|}{FiNER avg.}                 & \multicolumn{1}{l|}{42.60}           & \multicolumn{1}{l|}{37.78}          & 39.17          \\ \hline
\multicolumn{1}{|l|}{AttentionXML Pipeline avg.} & \multicolumn{1}{l|}{\tb{\textbf{45.51}}}  & \multicolumn{1}{l|}{\tb{\textbf{40.91}}} & \tb{\textbf{42.25}} \\ \hline
\end{tabular}}
\caption{\label{table:bucket analysis}Bucket analysis for benchmarked models}
\end{center}
\end{table}

% Further, with our trained model we performed inference with the data filed in the year-2019 (Table \ref{table:table_year_2019}) and year-2021 (Table \ref{table:table_year_2021}) respectively. 

% From the above analysis, it is established that the [NUM] and [SHAPE] pseudo-tokens help models successfully generalize over numeric expressions. With [SHAPE], we get best results in almost all scenarios. Numeric tokens of similar magnitudes may require similar xbrl tags. The shape pseudo-tokens capture each number’s magnitude.

\subsection{Hits@k}
Although we have evaluated based on exact match, the system may in practical setting 
%We identify that a major business use-case of this dataset is to use it to train systems that will 
recommend the top k tags to subject matter experts (SME) for a particular numeral which she may use to quickly produce the correct annotation. 
%The main principle is that when an annotator wants to label a particular numeral, the recommender would assist them by providing the top k labels quickly. Instead of selecting from hundreds of labels, the SME can simply inspect a short list of k top labels. 
We evaluate the \axml\ pipeline for this step and report the results in Table ~\ref{table:hits_at_k}. We observe that $\sim$90\% Hits@5 and $\sim$92\% Hits@10 is achieved for all the \axml\ pipeline models. This would mean that in more than 90\% of the cases the annotator would only have to inspect 5-10 labels. 

\begin{table}[ht]
\begin{center}
\resizebox{\columnwidth}{!}{
\begin{tabular}{|c|c|c|c|c|c|}
\hline
                                Model & Masking Token & Hits@1 & Hits@3 & Hits@5 & Hits@10      \\ \hline
\modelname\ & [MASK]   & 76.09 & \tb{\textbf{87.54}} & \tb{\textbf{90.14}} & \tb{\textbf{92.36}} \\ \hline
\modelname\ & [NUM]      & 75.89 & 87.24 & 89.96 & 92.06 \\ \hline
\modelname\ & [SHAPE]    & \tb{\textbf{76.76}} & 87.49 & 89.84 & 92.15 \\ \hline
\end{tabular}}
\caption{\label{table:hits_at_k}Hits@k results for \axml\ pipeline}
\end{center}
\end{table}

However, we need to inspect whether the task  really becomes easier. We first check the average cosine similarity between the top 5 labels predicted by the \axml\ pipeline and found it to be very high ( Figure ~\ref{fig:top_k_tag_sim}) which can easily confuse SMEs. We next carry out a human experiments with SMEs which is reported next. 
%\tb{We also observe that while [SHAPE] masking strategy performs best for Hits@1, the [MASK] masking strategy performs best for Hits@{3,5,10}.}

%\input{Sections/Analysis}
\subsection{Evaluation by Financial Domain Experts}

%To motivate the difficulty of our problem, we get 
We recruited a team of 6 financial SME who were asked to select the correct label for a numeral in a sentence from a list of top-5 ranked labels by the \axml\ [SHAPE] pipeline model. The experts are of Indian origin and non-native English speakers. They have been working in the industry for an average of 5 years (ranging from 2-10 years). The participants were remunerated equivalent to their half-a-day pay for their involvement in the task. The participants report that they could complete about 30 annotations per hour. It is worth noting that while the team members have an experience of 5 years in the industry on average, they have only a generic knowledge  on XBRL annotation.  
We ensure that the correct label was in the top 5 and this information was revealed to the participants.  In total, 305 datapoints were divided into 3 parts and each part was annotated by 2 SMEs. 
% \textcolor{green}{We filter the test cases for AttentionXML [SHAPE] pipeline model for the cases where the model is able to find the correct answer among the top 5 ranked. This participants were informed that the correct response would be a part of the options.}

%We find that in total, the annotators were collectively only able to correctly label 375 out of 610 data points (they felt no-confidence for the rest of the points). 

\begin{figure}[h]
    \centering
    \includegraphics[scale=0.35]{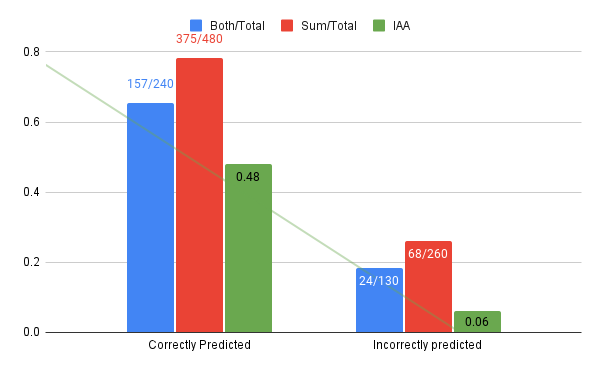}
    \caption{The three bar plots show (a). fraction of times both annotators predicted correctly (b). fraction of correct predictions and (c). inter-annotator agreement when \axml\ (i). predicted correctly and (ii). predicted incorrectly}
    \label{fig:human_performance}
    %\vspace{-6mm}
\end{figure}

The results, shown in Figure ~\ref{fig:human_performance}, show that  the annotators performed best when \axml\ has been able to predict correctly. 
Not only maximum labels were predicted correctly by either both or one of the annotators but the inter-annotator agreement was also high. 
The performance of humans deteriorated a lot for cases where machine has also failed. 

% A team of 6 subject matter experts (SME) were recruited for the evaluation task. 

%This calls for rigorous trainings for humans as well as design of a better human-AI interactive system. 
%Each data point is annotated by 2 participants and we find that only for 59.34\% of the data points were both participants able to correctly classify the datapoints. The overall Inter Annotator Agreement (IAA - $\kappa$ cohen) for these datapoints was 0.658. 

% \begin{figure}[h]
%     \centering
%     \includegraphics[scale=0.35]{Images/Human Performance hits@k.png}
%     \caption{Hits@k as labelled by annotators}
%     \label{fig:human_performance}
%     %\vspace{-6mm}
% \end{figure}

% \begin{figure}[h]
%     \centering
%     \includegraphics[scale=0.35]{Images/chart (1).png}
%     \caption{Hits@k as labelled by annotators}
%     \label{fig:human_performance}
%     %\vspace{-6mm}
% \end{figure}

% \begin{figure}[h]
%     \centering
%     \includegraphics[scale=0.45]{Images/scatter_random_vs_model.png}
%     \caption{Top-k tag similarity}
%     \label{fig:top_k_tag_sim}
%     %\vspace{-6mm}
% \end{figure}

% \begin{figure}[h]
%     \centering
%     \includegraphics[scale=0.45]{Images/scatter_st_2.png}
%     \caption{Top-k tag similarity \& human agreement}
%     \label{fig:top_k_tag_sim_agreement}
%     %\vspace{-6mm}
% \end{figure}

% \begin{figure}[h]
%     \centering
%     \includegraphics[scale=0.45]{Images/sorted_random_vs_model_agreemnt.png}
%     \caption{Top-k tag similarity \& human agreement}
%     \label{fig:sorted_top_k_tag_sim_agreement}
%     %\vspace{-6mm}
% \end{figure}
% \input{Sections/Discussion}
\section{Conclusion}
The paper provides a detailed idea about the challenges faced in tagging numerals with labels when the number of labels is large (2,794 tags)
and follow a long-tail distribution.
We have  rigorously collected an extensive set of labels, done an extensive bench-marking and executed a very specialized human experiment.
We believe the scope to include more information about the US-GAAP metrics label %GAAP metrics 
in the annotation model and a method to automate human-AI feedback loop would be the way forward to 
improve the performance of this difficult task. The dataset and codes are publicly available.

%XBRL tagging is a real-world time-consuming NLP task in the financial domain used for tagging business and financial reports to increase the transparency and accessibility of business information by using a uniform format. The human evaluation experiment clearly shows the difficulty of this task. To drive efforts towards automating this task, we released \dataset, a dataset of 79K sentences with 2794 xbrl tags. 
%Unlike typical entity extraction tasks, FNXL uses a much larger label set (2794 tags) where only financial numerals need to be tagged. We showcased that our proposed model could also provide reasonable performance in case of less frequent xbrl tags. Our higher hits@5 infers a potential business use case where the SME can simply inspect a short list of k top labels instead of selecting from thousands of XBRL tags and can achieve $\sim$90\% accurate result. Future work could exploit the hierarchical dependencies of XBRL tags and the short text description of tags.

\section{Limitations}
This work has only focussed on numerals from 10-K documents mandated by SEC. 
%However, we find that other documents such as 10-Q (quarterly reports) also contain XBRL annotations. We have not studied the information and the labels present in other documents. We also observe that there are other XBRL annotations provided in 10-K documents not limited to words and tabular structure. 
Our dataset, at present, does not include any annotated words as we focus only on numerals. It also does not include any tabular data. We also find that companies often annotate text with their custom labels which are not included in our dataset. We also find that often, it is difficult to label a numeral based on just the text of the sentence; the context might depend on surrounding paragraph, associated tables, etc. To this end, we have not benchmarked the performance using this information. However, we provide certain metadata along with the data points, including the company name, the year document was published, and the surrounding text which may be used to develop improved models.

\section{Ethics Statement}
Given the impact of our proposed contributions on the financial community in particular, and wider research community in general, our dataset and codes are publicly available. Our labels are derived from public/open domain. Still, we may ask users, intending to access our data, to provide a self declaration that the data is to be used solely for research purposes.

% While we release a dataset containing a large number of labels, we find that often it is difficult to label a numerical based on just the text of the sentence. To this end, we also include metadata such as surrounding paragraph text with a sentence, the company name and the year. We also find that sometimes the context for a sentence comes from a table associated with it. Our dataset, at present, does not include any tabular data. \soumya{Revisit} 

% Entries for the entire Anthology, followed by custom entries
\bibliography{acl2023}
\bibliographystyle{acl_natbib}

\newpage
\newpage
\appendix
\section{Appendix}~\label{sec:appendix}

\begin{figure*}[ht]
    \centering
    \includegraphics[scale=0.55]{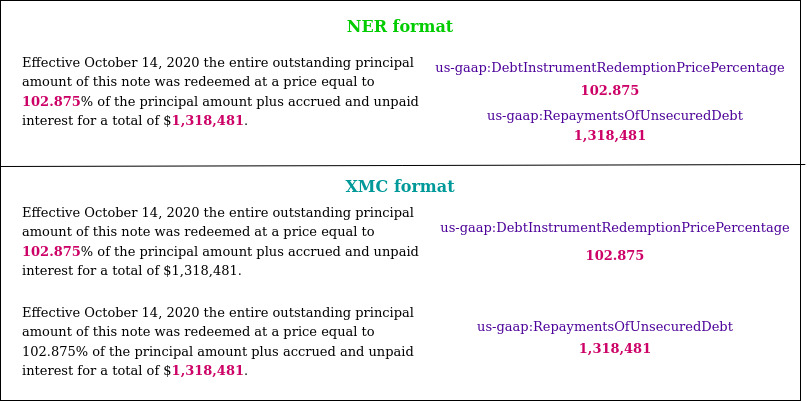}
    \caption{Dataset format}
    \label{fig:data_format}
    %\vspace{-6mm}
\end{figure*}

\subsection{Formats of the dataset}
We release two formats of the \dataset\ dataset. An example of this has been provided in Figure ~\ref{fig:data_format}

\begin{enumerate}
    \item The first format contains a sentence and all the associated numericals and its corresponding labels as NER tags.
    \item The second format contains a sentence and one of its corresponding numerical and its label. We structure the dataset in this format to adapt it to the Extreme Classification problem. In this format, we treat one numerical in a sentence as a single data point.
\end{enumerate}

\subsection{\axml\ pipeline}
\begin{figure}[ht]
    \centering
    \includegraphics[scale=0.4]{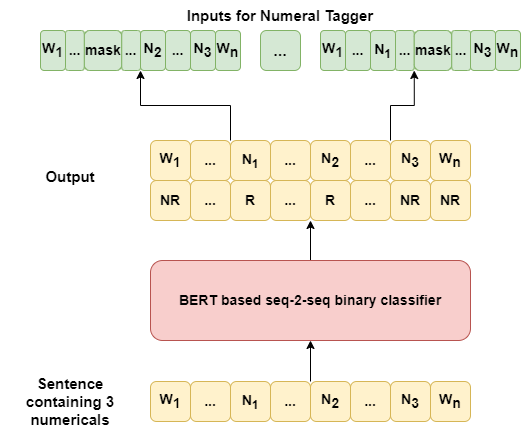}
    \caption{Binary Tagger output and processing of data to prepare input for AttentionXML. Here, N1, N2 and N3 represents 3 numericals out of which N1 and N2 are classified as relevant (R) and N3 is classified as non-relevant (NR) by the binary classifier. Depending on the number of relevant numericals, one data point is constructed by masking the non-focussed numericals.}
    \label{fig:pipeline}
    %\vspace{-6mm}
\end{figure}

In Figure ~\ref{fig:pipeline}, we showcase the pipeline method which uses \axml\ model. In this example, there are 3 numericals out of which 2 are classified as being relevant. For each relevant numerical, a new data point is created where the relevant numerical is left unmasked and the other numericals are masked using either the [MASK], [NUM] or [SHAPE] token.

\subsection{Model Hyperparameters}
For \axml\ model, we performed training for 30 epochs with batch size of 40, hidden size 256 and a dropout rate of 0.5. We trained binary tagger for 20 epochs with batch of 16 instances , learning rate 1e-5 and dropout of 0.1. For \finer\, we use a learning rate of 10**-4, 20 epochs, 32 batch size, 0.1 dropout rate. We use a single Tesla P100-PCIE (16GB) GPU. \axml\ model is trained in approximately 8 hours whereas \finer\ takes approximately 10-12 hours to train. 

\subsection{Bucket Analysis}

In Table ~\ref{table:table_top_100} and Table ~\ref{table:table_least_1000} we showcase the model performance for Top 100 frequently occurring labels and 1000 least frequently occurring labels.

\begin{table}[ht]
\begin{center}
\resizebox{\columnwidth}{!}{

\begin{tabular}{|c|c|c|c|c|}
\hline
Model & Masking Token & Macro-Precision & Macro-Recall & Macro-F1       \\ \hline
FiNER (bert-base) & no-masking               &   \textbf{91.33}    &   \textbf{79.22}    &   \textbf{83.79}    \\ \hline
FiNER (bert-base) & [NUM]           &   90.33    &   78.59    &    83.35   \\ \hline
FiNER (bert-base) & [SHAPE]         &   89.34    &    76.61   &     80.75  \\ \hline
FiNER avg. &  & \tb{\textbf{90.28}} &	\tb{\textbf{77.94}} &	\tb{\textbf{82.52}} \\ \hline\hline
% FiNER (sec-base)                &   90.91    &   80.36    &   84.90    \\ \hline
% FiNER (sec-num)                 &  91.13     &  79.30     &    83.95   \\ \hline
% FiNER (sec-shape)               &   90.70    &   76.8    &    82.00   \\ \hline
% \modelname + unmasked & 80.73 & 68.7  & 72.72 \\ \hline
\modelname & [MASK]   & 88.06 & 77.89 & 81.76 \\ \hline
\modelname  & [NUM]      & 88.86 & 77.25 & 81.5  \\ \hline
\modelname & [SHAPE]    & \textbf{89.50} & \textbf{78.45} & \textbf{82.62} \\ \hline
\modelname\ avg. &  & 88.81	& 77.87 &	81.97 \\ \hline
\end{tabular}}
\caption{\label{table:table_top_100}Top 100 frequent occurring classes}
\end{center}
\end{table}

\textbf{Top Frequently occurring labels:} We observe that \finer\ models perform better than \axml\ pipeline models. BERT-BASE \finer\ model with no masking achieves a 83.79\% Macro-F1 score whereas the best performing \axml\ pipeline model uses the [SHAPE] masking token and achieves a slightly lesser 82.62\%.

We also find that in the case of multi-numeral instances, \finer\ has a misclassification rate of 26.98\%, but \axml\ pipeline performs better with a misclassification rate of 28.94\%. While we also included zero-shot labels in the test and dev set, we find that both \finer\ and the \axml\ pipeline are not able to perform for zero-shot scenarios.

\begin{figure}[ht]
    %\centering
    \includegraphics[scale=0.45]{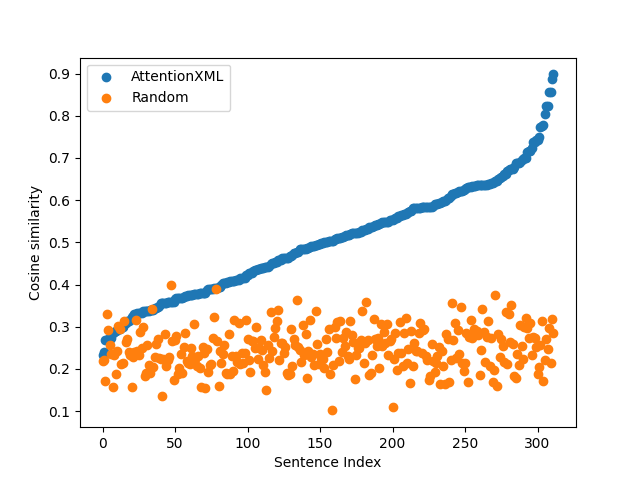}
    \caption{Top-k tag similarity}
    \label{fig:top_k_tag_sim}
    %\vspace{-6mm}
\end{figure}

\subsection{Binary Tagger vs \finer}
While the model structures for Binary Tagger in the \axml\ pipeline and \finer\ are the same, we observe that there are differences in their performance in tagging relevant numericals. We find that while \finer\ incorrectly tags a token in 1660 cases, the Binary Tagger only makes a mistake in $1406$ cases. We also observe that for $4$ examples, \finer\ incorrectly tagged a non-numerical as being relevant. For example, for the sentence "In March 2017, we issued \$350.0 million in aggregate principal amount of 1.25\% Senior Convertible Debentures due in 2025 (the '1.25\% 2025 Debentures') in a private placement.", while both models tag the numerical $350$ as relevant, \finer\ also identifies the two $1.25$ as being relevant. We attribute this difference in tagging between Binary Tagger and \finer\ to ease of task. While \finer\ has to decide between 2974 labels, Binary Tagger only has to make a binary decision, making the task easier for the model.

\subsection{Top-k tag similarity}

% \begin{figure}[h]
%     \centering
%     \includegraphics[scale=0.45]{Images/sorted_random_vs_model.png}
%     \caption{Top-k tag similarity}
%     \label{fig:top_k_tag_sim}
%     %\vspace{-6mm}
% \end{figure}

In Figure ~\ref{fig:top_k_tag_sim}, we showcase the average cosine similarity between the top 5 labels predicted by the \axml\ pipeline best performing model and compare it to the cosine similarity between the ground truth and randomly chosen 5 tags. We find that across all data points, the average similarity between the top 5 predicted tags is 0.5038 and the average similarity between the ground truth and randomly chosen 5 tags is 0.24. 
\section{Related Works}
%\textcolor{red}{Change style}
\textbf{Financial Resources and Applications:} While there has been a great deal of work on Natural language processing (NLP) for finance, it is still a relatively new field of study ~\cite{emnlp-2019-economics, finnlp-2020-financial, fnp-2020-joint}. There are few textual financial resources in the NLP literature. ~\citet{loukas2021edgar} published a corpus of all the US annual reports (10-K filings) from 1993-2020. ~\citet{handschke2018corpus} released JOCo, a corpus of non-SEC annual and social responsibility reports for the top 270 US, UK and German companies. ~\citet{daudert-ahmadi-2019-cofif} released CoFiF, the first financial corpus in the French language, comprising of annual, semestrial, trimestrial, and reference business documents. ~\citet{lee2014importance} released a collection of 8-K reports from EDGAR, which announce significant company events such as acquisitions or director resignations, from 2002-2012.

Financial documents have been used for a variety of tasks such as stock price prediction ~\cite{lee2014importance, chen2019group, yang2019leveraging}, risk analysis ~\cite{kogan2009predicting}, financial distress prediction ~\cite{gandhi2019using}, merger participants detection ~\cite{katsafados2021using}, financial relation extraction ~\cite{Sharma2022FinREDAD}, financial sentiment analysis ~\cite{malo2014good, Wang2013FinancialSA, akhtar2017multilayer}, summarization ~\cite{Rajdeep2022EMNLP}, economic event detection ~\cite{dor2019financial, jacobs2018economic, zhai2019forecasting} and causality analysis ~\cite{tabari2018causality, izumi2019economic, Nayak2022AGA}. %There has been a lot of research into understanding the structure of a financial report ~\cite{}. This is especially relevant for financial documents in the UK as there is no mandated structure for these reports.

\textbf{Entity Extraction:} XBRL tagging differs from NER task
and other previous entity extraction tasks (Table ~\ref{table:table_entity_stat} ).
In xbrl tagging there is a much
larger set of entity types (6k in full xbrl, 139 in finer-139, \dataset-2930) and the correct tag for numerical values highly depends on financial context. % For example, the maximum number of entitiy types we see in standard dataset such as conll-2003 ~\cite{tjong-kim-sang-de-meulder-2003-introduction}, ontonotes-v5 ~\cite{pradhan-etal-2013-towards},  

\begin{table}[ht]

\begin{center}
\resizebox{\columnwidth}{!}{
\begin{tabular}{|c|c|c|}
\hline
\textbf{Dataset}        & \textbf{Domain} & \textbf{Entity Types} \\ \hline
conll-2003              & Generic         & 4                     \\ \hline
ontonotes-v5            & Generic         & 18                    \\ \hline
ace-2005                & Generic         & 7                     \\ \hline
genia                   & Biomedical      & 36                    \\ \hline
~\citet{info10080248} & Financial       & 9                     \\ \hline
finer-139               & Financial       & 139                   \\ \hline
\dataset                    & Financial       & \textbf{2794}         \\ \hline
\end{tabular}}

\caption{\label{table:table_entity_stat}Examples of previous entity extraction
datasets}

\end{center}
\end{table}

% \subsection{NER}
% \subsection{XMC}
% \subsection{Large scale span labelling}
% \subsection{Financial Works}

% \section{Example Appendix}
% \label{sec:appendix}

% This is a section in the appendix.

\end{document}